\documentclass[letterpaper, 10 pt, conference]{ieeeconf}  

\IEEEoverridecommandlockouts                              

\usepackage{algorithm}
\usepackage{algpseudocode}
\usepackage{amsmath}
\usepackage{adjustbox}
\usepackage{multirow}

\usepackage{tikz}
\usepackage{textcomp}
\usepackage{hyperref}
\usepackage{lipsum}

\newcommand\copyrighttext{%
  \footnotesize \textcopyright 2025 IEEE. Personal use of this material is permitted.
  Permission from IEEE must be obtained for all other uses, in any current or future
  media, including reprinting/republishing this material for advertising or promotional
  purposes, creating new collective works, for resale or redistribution to servers or
  lists, or reuse of any copyrighted component of this work in other works.}
\newcommand\copyrightnotice{%
\begin{tikzpicture}[remember picture,overlay]
\node[anchor=south,yshift=10pt] at (current page.south) {\fbox{\parbox{\dimexpr\textwidth-\fboxsep-\fboxrule\relax}{\copyrighttext}}};
\end{tikzpicture}%
}

\title{\LARGE \bf
Adaptively Pruned Spiking Neural Networks for Energy-Efficient Intracortical Neural Decoding 
}

\author{Francesca Rivelli, Martin Popov, Charalampos S. Kouzinopoulos and Guangzhi Tang  
\thanks{Francesca Rivelli, Martin Popov, Charalampos S. Kouzinopoulos and Guangzhi Tang are with the Department of Advanced Computing Sciences, Faculty of Science and Engineering, Maastricht University, Maastricht, The Netherlands.
        {\tt\small guangzhi.tang@maastrichtuniversity.nl}}%
\thanks{This publication is part of the project \textit{Brain-inspired MatMul-free Deep Learning for Sustainable AI on Neuromorphic Processor} with file number NGF.1609.243.044 of the research programme AiNed XS Europe which is (partly) financed by the Dutch Research Council (NWO) under the grant https://doi.org/10.61686/MYMVX53467.}
}

\begin{document}

\maketitle
\copyrightnotice

\thispagestyle{empty}
\pagestyle{empty}

\begin{abstract}
Intracortical brain-machine interfaces demand low-latency, energy-efficient solutions for neural decoding. Spiking Neural Networks (SNNs) deployed on neuromorphic hardware have demonstrated remarkable efficiency in neural decoding by leveraging sparse binary activations and efficient spatiotemporal processing. However, reducing the computational cost of SNNs remains a critical challenge for developing ultra-efficient intracortical neural implants. In this work, we introduce a novel adaptive pruning algorithm specifically designed for SNNs with high activation sparsity, targeting intracortical neural decoding. Our method dynamically adjusts pruning decisions and employs a rollback mechanism to selectively eliminate redundant synaptic connections without compromising decoding accuracy. Experimental evaluation on the NeuroBench Non-Human Primate (NHP) Motor Prediction benchmark shows that our pruned network achieves performance comparable to dense networks, with a maximum tenfold improvement in efficiency. Moreover, hardware simulation on the neuromorphic processor reveals that the pruned network operates at sub-$\mu$W power levels, underscoring its potential for energy-constrained neural implants. These results underscore the promise of our approach for advancing energy-efficient intracortical brain-machine interfaces with low-overhead on-device intelligence.
\end{abstract}


\section{INTRODUCTION}

Intracortical neural implants play a pivotal role in brain-machine interfaces (BMIs) and neuroprosthetic systems by enabling the restoration of lost sensory, motor, or cognitive functions \cite{buccelli2019prosthesis, cervera2018rehabilitation}. For these devices to operate effectively, they must process neural signals in real-time within the brain under strict power constraints imposed by battery or wireless energy transfer \cite{lee2022miniaturized,miziev2024comparative}. Low-latency processing is essential to ensure immediate response, which is critical for applications such as prosthetic limb control and speech decoding \cite{willett2021high}. Furthermore, heat dissipation is a critical concern limiting the maximum power consumption, because even minor temperature increases in the sensitive environment of the human brain can jeopardize surrounding tissues \cite{serrano2020thermal}. Recent advances in deep learning have significantly improved the accuracy of intracortical neural decoding \cite{liu2022deep,mathis2024decoding}. However, the substantial computational cost associated with deep neural networks hinders their direct implementation in resource-constrained intracortical neural implants, thereby necessitating the development of more efficient solutions.

Neuromorphic computing using Spiking Neural Networks (SNNs) has demonstrated significant energy efficiency improvements across a broad spectrum of applications \cite{yin2021accurate,paredes2024fully,kumar2022decoding,xu2025event}. In contrast to conventional deep neural networks, SNNs process spatiotemporal information efficiently by leveraging sparse, event-driven computations and stateful spiking neurons \cite{tavanaei2019deep}. SNN-based approaches for intracortical neural decoding have been investigated and have demonstrated performance comparable to state-of-the-art methods, while incurring lower energy costs \cite{hueber2024benchmarking}. Furthermore, the collaborative NeuroBench benchmark for neuromorphic computing has designated intracortical neural decoding as one of its initial tasks, underscoring the significance of this application in demonstrating the advantages of neuromorphic systems \cite{yik2025neurobenchframeworkbenchmarkingneuromorphic}. Although SNNs have already demonstrated efficient neural decoding, an open challenge remains in further reducing computational costs to meet the stringent power constraints, ultimately at the sub-$\mu$W level \cite{chatterjee2023biphasic}, required for ultra-efficient intracortical neural implants.

Synaptic pruning reduces the computational cost of neural networks by eliminating unnecessary neurons and synaptic connections \cite{cheng2024survey}. Various pruning strategies employ metrics such as synaptic weight magnitudes \cite{frankle2018lottery}, changes in loss \cite{hassibi1993optimal}, and activation patterns \cite{he2023structured} to systematically remove network components that contribute minimally to overall performance. In SNNs, pruning not only reduces the number of weights but also decreases the number of synaptic operations per inference, thereby lowering overall computational costs. For instance, weight magnitude-based pruning \cite{gupta2024pruning} and activity-based pruning \cite{li2022spikingpruning} are two methods that target distinct SNN characteristics to enhance efficiency. However, additional challenges arise when SNNs exhibit extremely high activation sparsity during neural decoding. In such cases, the network becomes highly sensitive to synaptic pruning, and conventional pruning methods can lead to significant performance degradation.

In this paper, we propose an adaptive pruning algorithm to further enhance the energy efficiency of SNNs for intracortical neural decoding. Our algorithm accelerates both the pruning speed and overall pruning rate by incorporating an adaptive execution strategy, and it directly addresses the sensitive pruning problem through an adaptive rollback mechanism during the pruning process. We evaluated our approach using the NeuroBench Non-Human Primate (NHP) Motor Prediction benchmark \cite{yik2025neurobenchframeworkbenchmarkingneuromorphic}. Compared with dense networks of similar architectures, SNNs pruned by our adaptive pruning algorithm achieve comparable prediction performance while delivering over a maximum 10$\times$ improvement in efficiency. Furthermore, a hardware simulation study employing realistic measurements from the SENECA neuromorphic processor \cite{xu2024optimizing} demonstrates that our pruned SNN can achieve sub-$\mu$W power consumption, thereby significantly reducing the computational overhead of neural decoding in intracortical neural implants.

\section{Method}

We employ Spiking Neural Networks (SNNs) with stateful Leaky Integrate-and-Fire (LIF) neural models to decode spatiotemporal information from intracortical neural signals recorded during the Primate Reaching task. To improve energy efficiency and reduce computational and memory requirements, we developed an adaptive pruning algorithm that minimizes the number of synapses and synaptic operations without compromising the performance of the SNN.

\subsection{Spiking Neural Network}

\begin{figure} [h]
\includegraphics[width=0.5\textwidth]{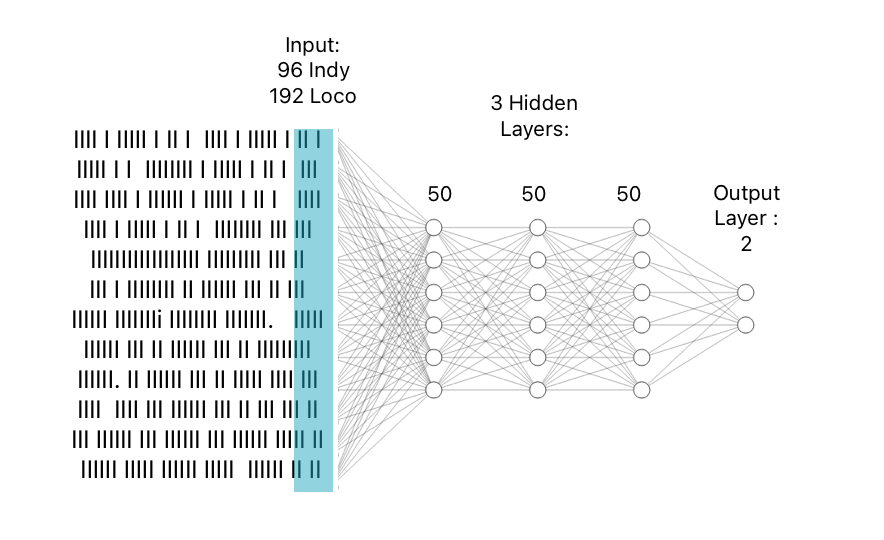}
\centering
\caption{Spiking neural network with 3 hidden layers for decoding recorded neural signals of the Primate Reaching task. The input, hidden, and output dimensions are listed. The adaptive pruning algorithm is applied to the hidden layers of the network.}
\label{fig:img_ntwk}
\end{figure}

In this study, we employ a multilayer SNN, illustrated in Figure~\ref{fig:img_ntwk}, to decode neural signals recorded during the Primate Reaching task. The architecture, referred to as SNN3 in \cite{hueber2024benchmarking}, comprises three hidden layers and one output layer. All hidden layers consist of spiking LIF neurons, while the output layer comprises two non-spiking LIF neurons whose membrane potentials encode the two-dimensional finger velocity. The inputs are binary spikes from thresholding the intracortical neural recordings. The input dimension varies by session, with 96 channels for Indy sessions and 192 channels for Loco sessions, reflecting differences in the recording devices. The fully connected SNN contains 9,900 synapses for Indy sessions and 14,700 synapses for Loco sessions. We applied our adaptive pruning algorithm to reduce the number of synapses (weights) in the hidden layers of the network.

\subsection{Leaky Integrate-and-Fire (LIF) Neurons}

In our SNN, stateful LIF neurons are employed to capture the spatiotemporal dynamics of the recorded neural signals. At each timestep, each LIF neuron updates its state through two sequential processes \cite{wu2018spatio}: first, the membrane potential is updated, and then the neuron may generate a spike. In the hidden layers, neurons execute both the membrane potential update and the spiking process, whereas neurons in the output layer update their membrane potentials without generating spikes.

\subsubsection{Membrane potential update}

The membrane potential of a neuron at a given time \( t \), denoted as \( u(t) \), is influenced by its previous potential \( u(t_{i-1}) \) at time \( t_{i-1} \) and the general pre-synaptic input \( \hat{I}(t) \). The equation governing this process is given by:

\begin{equation}
    u(t) = u(t_{i-1}) e^{\frac{t_{i-1} - t}{\tau}} + \hat{I}(t)
\end{equation}

In this formulation, the term \(  e^{\frac{t_{i-1} - t}{\tau}} \) represents the exponential decay of the previous potential over time, where \( e \) acts as the exponential decay factor, and \( \tau \) is the time constant that determines the rate of decay. The second term, \( \hat{I}(t) \), accounts for the effect of synaptic input received at the current time step, which is the weighted sum of input spikes. This equation describes how the neuron’s potential evolves over time based on both its past state and incoming synaptic activity.

\subsubsection{Spiking and reset}

A neuron generates a spike when its membrane potential \( u(t) \) reaches or exceeds a predefined threshold \( \theta \). Mathematically, this condition is expressed as:

\begin{equation}
    u(t) \geq \theta
\end{equation}

where \( u(t) \) represents the neuron's membrane potential, and \( \theta \) is the spiking threshold that determines when an action potential is triggered. Our threshold in the SNN is set to one.

After a spike occurs, the membrane potential is reset to a lower value, denoted as \( u_{\text{reset}} \), to ensure proper neuronal dynamics. This reset mechanism is described by:

\begin{equation}
    u(t) \leftarrow u_{\text{reset}}
\end{equation}

where \( u_{\text{reset}} \) is the reset potential that the neuron returns to immediately after firing a spike. We set the reset potential to zero in our SNN. This mechanism prevents continuous firing and allows the neuron to undergo a refractory period before it can spike again.

\subsection{Adaptive Pruning for Spiking Neural Network}

\begin{figure}[b]
\includegraphics[width=8cm]{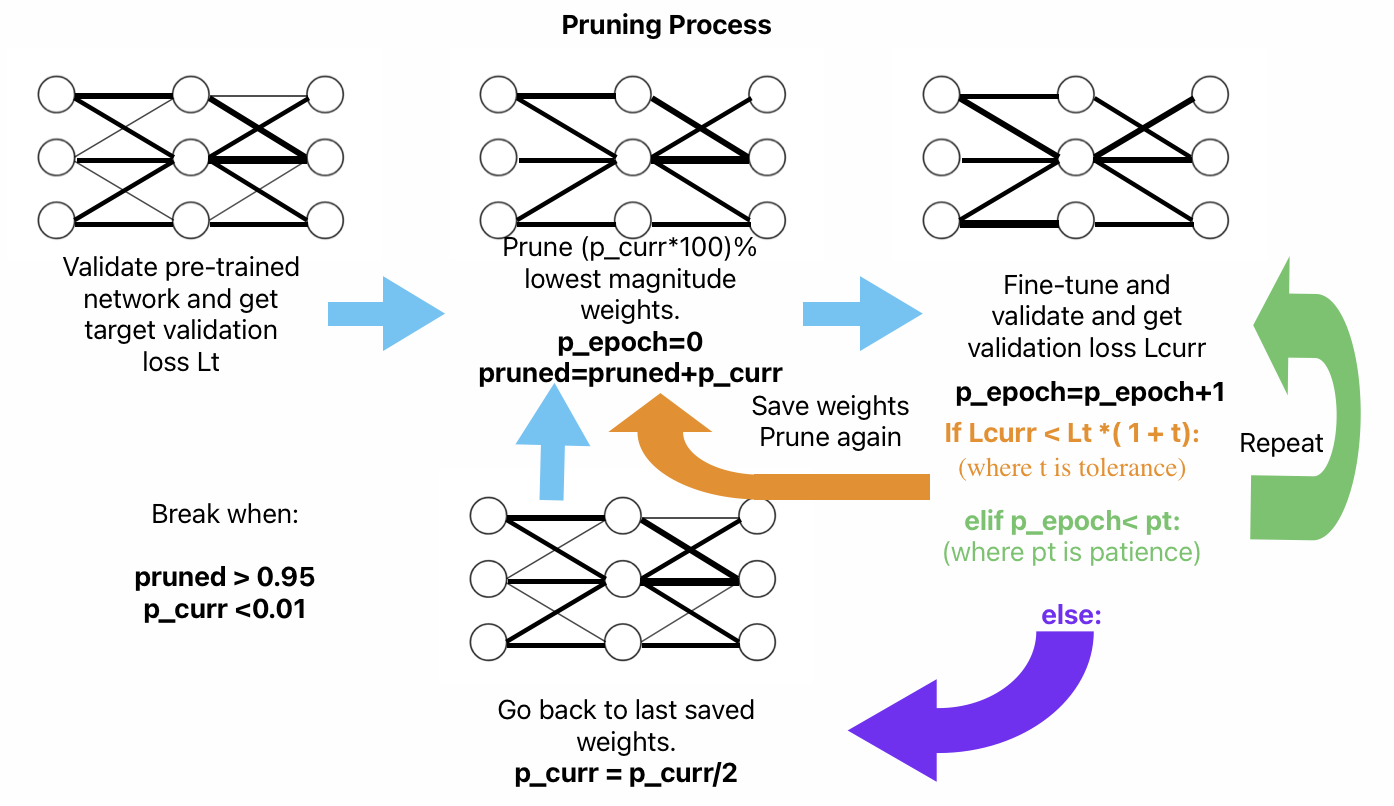}
\centering
\caption{General steps of our proposed adaptive pruning algorithm for SNN. The pruning process keeps iterating until either the target pruned connection percentage (pruned) is reached or the adaptive pruning rate reaches the minimal value (pr).}
\label{fig:pruning_pipeline}
\end{figure}

We propose the adaptive pruning algorithm to prune the synaptic connections of SNNs. The general steps of the algorithm are presented in Figure~\ref{fig:pruning_pipeline}. Compared to regular pruning methods, our approach adapts the pruning rate, fine-tuning interactions, and pruned model based on the ongoing pruning performance determined by the validation dataset. This adaptive behavior reduces the number of training iterations during pruning, and improves the sparsity of the model while maintaining the performance.

We begin the pruning process with a dense SNN pre-trained on the training dataset. The validation loss of this pre-trained network, denoted by $L_t$, is used as the target loss throughout the pruning process. Every pruning iteration prunes the $p\%$ lowest magnitude weights and is followed by one or more fine-tuning epochs using the training dataset. Initially, the \textbf{Starting Pruning Rate} is set at its maximum value, as the network is more robust to pruning when it contains many synapses. After the first iteration, pruning is performed only after the pruned SNN has been fine-tuned to achieve the target loss. Recognizing that the pruned network may not exactly match $L_t$, we introduce a \textbf{Pruning Tolerance} range around $L_t$ within which the loss is considered acceptable for further pruning. Additionally, the number of fine-tuning epochs is limited by the \textbf{Pruning Patience} parameter. If the network fails to achieve the target loss within the granted epochs, the current pruning rate is deemed too aggressive. In such cases, the most recently pruned model is discarded and the pruning rate is halved. A detailed description of the adaptive pruning algorithm is provided in Algorithm~\ref{alg}. Three major hyperparameters are introduced to control the adaptiveness of our algorithm:

\begin{enumerate}
    \item \textbf{Starting Pruning Rate}: Pruning rate for the first pruning iteration and representing the highest pruning rate during the pruning process. The rate influences the pruning speed by directly influencing the pruning magnitude at each step.
    \item \textbf{Pruning Tolerance}: The rate of the pre-trained validation loss that the validation loss of pruned networks is "tolerated" to exceed for executing pruning. This is introduced to take care of random fluctuations of loss during training.
    \item \textbf{Pruning Patience}: The number of fine-tuning epochs to wait before halving the pruning rate and discarding the most recent pruned network.
\end{enumerate}

\begin{algorithm}[t]

\caption{SNN Adaptive Pruning Algorithm}
\begin{algorithmic}[1]

\State Minimum pruning rate $p_{min} \gets 0.1$ and maximum pruning possible $ pruned_{max} \gets 0.95 $ are fixed. 
\State \textbf{Initialize:}  
\State \hspace{1em} Starting Pruning Rate: $p_{start}$  
\State \hspace{1em} Pruning Patience: $pt$  
\State \hspace{1em} Pruning Tolerance: $t$  
\State \hspace{1em} \% of network that has been pruned:  $pruned$
\State Validate the network to obtain the initial target loss $L_{t}$  
\State Set the current pruning rate: $p_{curr} = p_{start}$  

\While{$p_{curr} \geq p_{min}$ \textbf{and} $pruned < pruned_{max} $}
    \State Apply pruning with rate $p_{curr}$  
    \State Update: $pruned = pruned + p_{curr}$  
    \State Initialize current loss: $L_{curr} \gets \infty$  
    \State Reset fine-tuning epoch counter: $p_{epoch} \gets0$  

    \While{$L_{curr} > L_{t} (1 + t)$}
        \If{$p_{epoch} > pt$}
            \State Restore previously pruned SNN  
            \State Reduce pruning rate: $p_{curr} \gets p_{curr} /2$  
            \State \textbf{break}
        \EndIf
        \State Fine-tune epoch of the pruned SNN  
        \State Compute validation loss: $L_{curr}$  
        \State $p_{epoch} \gets p_{epoch} + 1$  
    \EndWhile  
\EndWhile  

\end{algorithmic}
\label{alg}
\end{algorithm}

Our synaptic pruning algorithm can be performed either globally or on a per-layer basis on the SNN. In the global approach, the $p\%$ lowest-magnitude weights across all hidden layers of the SNN are pruned, whereas in the per-layer approach, the $p\%$ lowest-magnitude weights in each hidden layer are pruned. We adopt the per-layer pruning approach, as it yields a more balanced pruned SNN compared to the global method. During GPU-based training, pruning is implemented via element-wise multiplication of the weight matrix with a binary mask, where pruned weights are assigned a value of 0 and retained weights a value of 1. We adopt a static pruning approach in which the same mask is reapplied at every batch during fine-tuning and prior to validation. This ensures that pruned weights do not regain relevance over time.

\section{Experiments and Results}

\subsection{Datasets and Experiments}

We benchmarked the performance of our pruned SNN using the Primate Reaching task from the neuromorphic benchmark NeuroBench \cite{yik2025neurobenchframeworkbenchmarkingneuromorphic}. The task comprises data from 6 recording sessions obtained from 2 Rhesus primates (Indy and Loco), with each subject contributing 3 sessions \cite{makin2018superior}. According to the NeuroBench benchmark, each session is subdivided into 4 sub-sessions, within each sub-session 50\% of the data is allocated for training, 25\% for validation, and 25\% for testing. The task output is a two-dimensional prediction of finger velocity, represented by the X and Y coordinates.

Since the NeuroBench benchmark uses a different split for the training, validation, and test datasets than the evaluation presented in \cite{hueber2024benchmarking}, we retrained the dense SNN3 model using the NeuroBench split so that it could serve as the pre-trained network for pruning. We applied adaptive pruning to the pre-trained models from all 6 sessions using identical hyperparameters for $Starting\ Pruning\ Rate=10$ and $Pruning\ Patience=5$, and different $Pruning\ Tolerance=0.1/0.05$ based on the model validation performance. Furthermore, we conducted experiments to compare various pruning approaches and performed ablation studies using data from the first recording session in the dataset.

\subsection{NeuroBench Harness and Selected Metrics}

\begin{table*}
\caption{Performance comparison between baseline dense networks and our pruned networks.}
  \label{tb:main-result}
  \begin{center}
\begin{tabular}{ccccccccc}
\hline
 \multicolumn{1}{c|}{} & \multicolumn{4}{c|}{Average of Indy sessions} & \multicolumn{4}{c}{Average of Loco sessions}\\ 
\hline
 \multicolumn{1}{c|}{Method} & $R^2$ & Connection & Activation & \multicolumn{1}{c|}{Effective} & $R^2$ & Connection & Activation & Effective\\
 \multicolumn{1}{c|}{} & & Sparsity & Sparsity & \multicolumn{1}{c|}{Operations} & & Sparsity & Sparsity & Operations\\
\hline
\multicolumn{1}{c|}{NeuroBench ANN \cite{yik2025neurobenchframeworkbenchmarkingneuromorphic}} & 0.593 & 0.0 & 0.683 & \multicolumn{1}{c|}{3836 MACs} & 0.558 & 0.0 & 0.668 & 6103 MACs\\
\multicolumn{1}{c|}{NeuroBench SNN \cite{yik2025neurobenchframeworkbenchmarkingneuromorphic}} & 0.593 & 0.0 & 0.997 & \multicolumn{1}{c|}{276 ACs} & 0.568 & 0.0 & 0.999 & 551 ACs\\
\multicolumn{1}{c|}{Dense SNN3 \cite{hueber2024benchmarking}} & 0.583 & 0.0  & 0.9813  & \multicolumn{1}{c|}{408.14 ACs} & 0.570  & 0.0  & 0.9893  & 628.10 ACs\\
\hline
\multicolumn{1}{c|}{\textbf{Our adaptive pruning}+SNN3} & 0.570  & \textbf{0.8915}  & 0.9826  & \multicolumn{1}{c|}{\textbf{38.37 ACs}} & 0.546  & \textbf{0.791}  & 0.9898  & \textbf{95.38 ACs}\\
\hline
\end{tabular}
  \end{center}
\end{table*}

To fairly compare with other dense networks, we tested all our pruned SNNs using the NeuroBench Harness library. We selected the following metrics from NeuroBench that are relevant to our experiments:
\begin{enumerate}
    \item \textbf{R² (coefficient of determination)}: This metric is calculated separately for the \( X \)- and \( Y \)-velocity components and taking the average. To determine the correctness score for a session, the \( R^2 \) values for both velocity components are averaged.
    \begin{equation}
        R^2 = 1 - \frac{\sum\limits_{i=1}^{n} (y_i - \hat{y}_i)^2}{\sum\limits_{i=1}^{n} (y_i - \bar{y})^2}
    \end{equation}
    where $n$ is the number of labeled points in the test dataset, $y_i$ is the groundtruth velocity at each point, the predicted velocity as $\hat{y}_i$ the predicted velocity and $\bar{y}$ the mean of groundtruth velocities.
    \item \textbf{Connection Sparsity}:
    Connection sparsity represents the percentage of zero-value synaptic weights in a neural network. It is computed by dividing the number of zero-value synaptic weights by the overall number of weights in the network.
    \item \textbf{Activation Sparsity}: 
    The average sparsity of neuron activations across all layers, timesteps, and input samples during network execution. A value of 0 indicates that all neurons generate non-zero activation values, while 1 means all neurons generate activation values equal to zero.
    \item \textbf{Effective Synaptic Operations}: The average number of effective synaptic operations during network execution, determined by non-zero neural activations and synaptic weights. This metric is categorized into Multiply-and-Accumulate (MAC) operations for ANNs, and Accumulate (AC) operations for SNNs.
\end{enumerate}

Detailed descriptions of each selected metric can be found in the NeuroBench Harness \cite{yik2025neurobenchframeworkbenchmarkingneuromorphic}.

\subsection{Comparisons with Baseline Dense Networks}

We evaluated the test performance of our adaptive pruning algorithm relative to dense networks using the selected Neurobench metrics and the benchmarking harness. Table~\ref{tb:main-result} presents the average results across different recording sessions. Compared to dense SNN and ANN approaches, our sparsely pruned SNN achieves comparable performance while using only 10\% of the effective synaptic operations in most sessions. With similar activation sparsity, this drastic reduction in synaptic operations is entirely attributable to the increased connection sparsity introduced by our pruning algorithm. Since the network operation cost on hardware is proportional to the number of synaptic operations, this reduction directly enhances the energy efficiency of network inference.

Although our pruned network achieves performance comparable to that of dense networks, pruning still results in a reduction in $R^2$. To investigate the cause of this performance reduction, Table~\ref{tb:main-result-per-session} presents a detailed comparison of the performance between the pre-trained dense SNN and the pruned SNN for each recording session. We observed that the reduction in $R^2$ varies across recording sessions. Specifically, $R^2$ remains relatively stable for \textit{indy\_20160622\_01}, \textit{indy\_20170131\_02}, and \textit{loco\_20170301\_05}, whereas other sessions exhibit more pronounced declines after pruning. This suggests that the pruning process may be overly aggressive for these sessions, likely due to their already extremely high activation sparsity (exceeding 99\%), which increases sensitivity to synaptic pruning. To mitigate this effect, we reduced the pruning tolerance for the unstable sessions.


    



\begin{table*}
\caption{Performance comparison of each recording session between dense networks and our pruned networks.}
\label{tb:main-result-per-session}
\begin{center}
    \begin{tabular}{ccc|ccccc}
    \hline
    Session name & Method & Tolerance & $R^2$ & Connection & Activation & Effective & Pruned \\
     & & & & Sparsity & Sparsity & Operations (ACs) & Rate (\%)\\
    \hline
    \multirow{2}{*}{indy\_20160622\_01} & Dense SNN3\cite{hueber2024benchmarking} & - & 0.6618 & 0.0 & 0.9789 & 535.2 & -\\
    & \textbf{Adaptive Pruning} & 0.1 & 0.6539 & 0.8978 & 0.9763 & 54.63 & 90\\
    \hline
    \multirow{2}{*}{indy\_20170131\_02} & Dense SNN3\cite{hueber2024benchmarking} & - & 0.5812 & 0.0 & 0.9711 & 391.0 & -\\
    & \textbf{Adaptive Pruning} & 0.1 & 0.5653 & 0.9072 & 0.9786 & 27.88 & 89\\
    \hline
    \multirow{3}{*}{indy\_20160630\_01} & Dense SNN3\cite{hueber2024benchmarking} & - & 0.5065 & 0.0 & 0.9937 & 298.1 & -\\
    & \textbf{Adaptive Pruning} & 0.1 & 0.4559 & 0.7966 & 0.9930 & 45.67 & 80\\
    & \textbf{Adaptive Pruning} & 0.05 & 0.4919 & 0.8696 & 0.9929 & 32.60 & 87\\
    \hline
    \multirow{2}{*}{loco\_20170301\_05} & Dense SNN3\cite{hueber2024benchmarking} & - & 0.5978 & 0.0 & 0.9852 & 720.3 & -\\
    & \textbf{Adaptive Pruning} & 0.1 & 0.5805 & 0.9062 & 0.9890 & 50.91 & 91\\
    \hline
    \multirow{2}{*}{loco\_20170215\_02} & Dense SNN3\cite{hueber2024benchmarking} & - & 0.5475 & 0.0 & 0.9919 & 608.0 & -\\
    & \textbf{Adaptive Pruning} & 0.1 & 0.4859 & 0.8810 & 0.9883 & 48.57 & 88\\
    & \textbf{Adaptive Pruning} & 0.05 & 0.5145 & 0.6678 & 0.9911 & 155.85 & 66\\
    \hline
    \multirow{2}{*}{loco\_20170210\_03} & Dense SNN3\cite{hueber2024benchmarking} & - & 0.5648 & 0.0 & 0.9906 & 555.8 & -\\
    & \textbf{Adaptive Pruning} & 0.1 & 0.5295 & 0.8714 & 0.9884 & 45.60 & 87\\
    & \textbf{Adaptive Pruning} & 0.05 & 0.5424 & 0.7991 & 0.9893 & 79.37 & 79\\
    \hline
    \end{tabular}
\end{center}
\end{table*}

\subsection{Comparing Global and Per-layer Pruning}

Table~\ref{tb:global-vs-layer} compares the single session performance of global and per-layer pruning approaches. Our results indicate that while the global pruning method achieves performance similar to that of the per-layer approach, it exhibits a slightly lower $R^2$. The per-layer pruning approach is preferred because it enforces a uniform pruning rate across all layers, resulting in similar connection sparsity within each layer. This uniformity benefits network deployment on neuromorphic hardware, as it ensures that each layer consumes similar computational resources and operates within comparable time constraints.

\begin{figure}[!htpb]
\includegraphics[width=8cm]{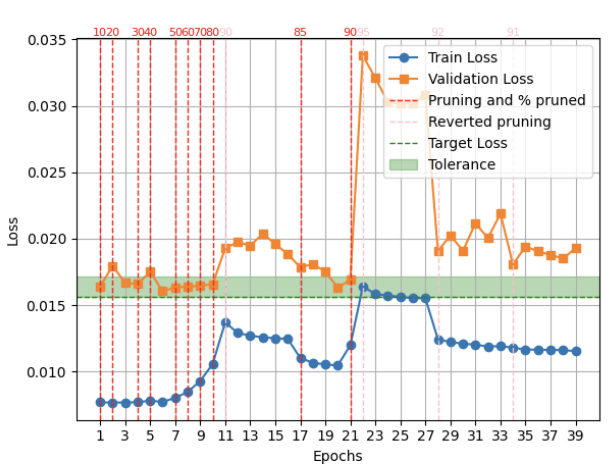}
\centering
\caption{Training and validation loss during the adaptive pruning process using the \textbf{per-layer} approach.}
\label{fig:pruning_loss_l}
\end{figure}

\begin{figure}[!htpb]
\includegraphics[width=8cm]{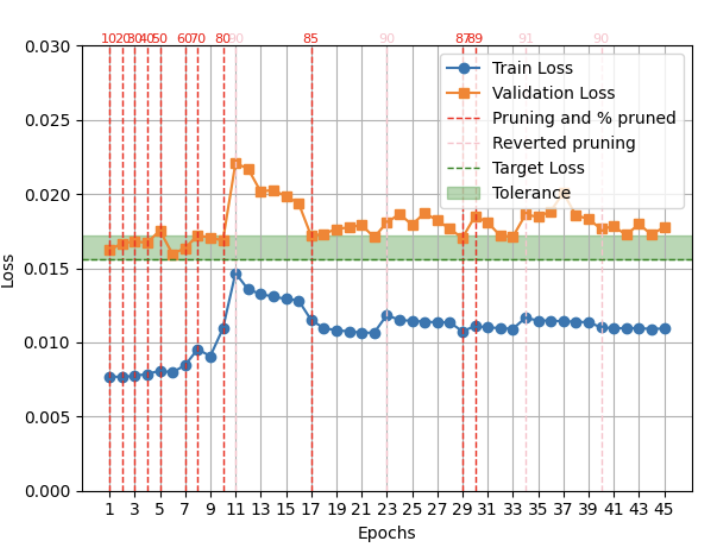}
\centering
\caption{Training and validation loss during the adaptive pruning process using the \textbf{global} approach.}
\label{fig:pruning_loss_g}
\end{figure}

Figures~\ref{fig:pruning_loss_l} and~\ref{fig:pruning_loss_g} show the detailed adaptive pruning process for both per-layer and global approaches on a single recording session. In the figures, the red dotted lines indicate the epochs at which pruning operations are executed, with the network pruned rate after pruning annotated above each line. The shaded lines represent pruning operations that were ultimately discarded because the network failed to reach the target loss after fine-tuning. The pruning process terminates once the exit criteria are met. In both approaches, the process terminates upon reaching the minimum pruning rate. This minimum is achieved after multiple adaptive reductions in the pruning rate, prompted by the network's failure to achieve the target loss after fine-tuning. Consequently, the total number of fine-tuning epochs exceeds the optimal effective epochs required to obtain the final pruned SNN. However, we argue that these additional training epochs are not wasted, as they represent an exploration process that facilitates further improvements in sparsity.

\begin{table}
    \caption{Comparison of global and layer-wise pruning approaches.}
\label{tb:global-vs-layer}
\begin{center}
    \begin{tabular}{ccc}
    \hline
    Metric & Global Pruning & Layer Pruning\\
    \hline
    $R^2$ & 0.6333 & 0.6539 \\
    Connection Sparsity & 0.8811 & 0.8978 \\
    Activation Sparsity & 0.9724 & 0.9763 \\
    Effective ACs & 57.61 & 54.63 \\
    Pruned Rate (\%) & 89\% & 90\% \\
    Total Epoch & 45 & 39 \\
    Optimal Epoch & 33 & 21 \\
    \hline
    \end{tabular}
\end{center}
\end{table}

\subsection{Ablation Studies on Adaptive Pruning}

\begin{table*}
\caption{Performance comparison of different ablation studies.}
\label{tb:ablation}
\begin{center}
    \begin{tabular}{cc|ccccccc}
    \hline
    \multicolumn{2}{c|}{Ablations} & $R^2$ & Connection & Activation & Effective & Pruned & Total & Optimal \\
     Tolerance & Rate Decay & & Sparsity & Sparsity & Operations (ACs) & Rate (\%) & Epoch & Epoch\\
    \hline
    No & No & 0.515 & 0.902 & 0.975 & 55.76 & 90\% & 40 & -\\
    Yes & No & 0.661 & 0.812 & 0.977 & 102.88 & 80\% & 16 & 10\\
    Yes & Yes & 0.654 & 0.898 & 0.976 & 54.63 & 90\% & 33 & 21\\
    \hline
    \end{tabular}
\end{center}
\end{table*}

We conducted two ablation studies to evaluate the contributions of individual components in our adaptive pruning algorithm. Specifically, we investigated the roles of the tolerance and rate decay components. The tolerance component governs when the network initiates pruning based on the validation loss, while the rate decay component discards ineffectively pruned models and reduces the pruning rate over time. In the first study, we implement fixed pruning by omitting both the tolerance and rate decay components. In this scheme, 10\% of the network is pruned after every 5 epochs of fine-tuning, and the process continues until the overall pruning rate reaches 90\%. In the second study, we incorporate the tolerance component, enabling the pruning process to decide when to prune based on the validation loss. However, this study omits rate decay. The pruning process terminates when the target loss cannot be achieved within the number of fine-tuning epochs specified by \textbf{Pruning Patience}.

Table~\ref{tb:ablation} presents the performance comparison between the ablation studies and the full adaptive pruning algorithm on a single recording session. Compared to the fixed pruning approach, our adaptive pruning method achieves higher $R^2$ while maintaining the same connection sparsity and requiring fewer training epochs. These results demonstrate the effectiveness of the adaptive approach in achieving rapid pruning and high performance. Furthermore, compared to a variant that employs only the tolerance component, the full adaptive algorithm achieves higher connection sparsity and reduces the effective number of synaptic operations by nearly 50\%, while maintaining similar $R^2$. These findings indicate that incorporating rate decay and adaptive rollback on aggressively pruned networks enables the pruning process to achieve higher sparsity without sacrificing performance.

    

\subsection{Hardware Simulation on Neuromorphic Processor}

To evaluate the hardware benefits of our adaptive pruning algorithm, we conducted a hardware simulation study using realistic measurements obtained from the SENECA neuromorphic processor \cite{tang2023open}. SENECA is an energy-efficient, flexible digital neuromorphic processor designed for sparse, event-driven neural networks \cite{xu2024optimizing}. To implement our sparse SNN on SENECA, we programmed two micro-kernels to handle spike integration and membrane potential updates for LIF neurons. On SENECA, integrating a single spike (1 AC) into a LIF neuron incurs an energy cost of $12.7\ \text{pJ}$, while updating a LIF neuron at each timestep requires $14.6\ \text{pJ}$. Table~\ref{tb:hw} presents the hardware simulation results for one timestep of both the dense and pruned SNNs based on benchmarking results from one recording session. The SNN inference timestep corresponds to $4\ \text{ms}$ of neural decoding, and this time bin is used to compute the average power consumption of the SNN. The results demonstrate that our adaptive pruning approach significantly reduces the energy cost of neural decoding and requires minimal power that can potentially support intracortical decoding.

\begin{table}[h]
\caption{Neuromorphic Hardware Simulation Comparison.}
\label{tb:hw}
\begin{center}
    \begin{tabular}{c|cccc}
    \hline
    Method & Energy Cost (pJ) & Power ($\mu$W)\\
    \hline
    Dense SNN & 6811.6 & 1.7\\
    Pruned SNN & 708.4 & 0.18\\
\hline
    \end{tabular}
\end{center}
\end{table}

\section{Conclusion and Discussion}

In this work, we introduce an adaptive pruning algorithm designed to reduce synaptic operations in SNNs for intracortical neural decoding. Our experimental findings indicate that this strategy effectively increases connection sparsity, thereby lowering computational demands and enhancing energy efficiency in brain-machine interfaces. By selectively eliminating redundant synaptic connections, our approach optimizes performance without compromising decoding accuracy, paving the way for more energy-efficient neural implant solutions.

Our adaptive pruning algorithm is specifically designed for SNNs that exhibit high activation sparsity, a property that makes them particularly harmed by performance degradation during pruning. While our approach has been demonstrated on a baseline SNN architecture, its modular design facilitates extension to more complex recurrent SNN models for neural decoding \cite{liu2024decoding}. Adapting the algorithm for recurrent networks will require a balancing strategy that distributes pruning between recurrent and feedforward connections. Given that SNNs can effectively perform spatiotemporal processing even in the absence of recurrent connections, it is possible that prioritizing the pruning of recurrent connections could be achieved without compromising overall performance.

Our adaptive pruning algorithm can be synergistically integrated with complementary network compression techniques, such as quantization \cite{liang2021pruning}, to further enhance the efficiency of intracortical neural decoding. Applying quantization to the remaining synaptic weights not only reduces memory requirements but also lowers the energy cost per synaptic operation. Dedicated neuromorphic hardware is essential to fully leverage the efficiency gains of the pruned SNN \cite{chen2024sibrain,kuang2022essa}. However, current neuromorphic platforms may not adequately address the unique requirements of intracortical neural implants \cite{wang2023implantable}. Hence, further hardware development is necessary to realize the full benefits of our approach in practical applications.

\bibliographystyle{IEEEtran}
\bibliography{bibliography}

\end{document}